\documentclass{edm_article}
\usepackage{soul}
\usepackage{svg}
\usepackage{graphicx}
\usepackage{multirow}
\usepackage{placeins}
\begin{document}



\title{Integrating Cognitive AI with Generative Models for Enhanced Question Answering in Skill-based Learning}
\author{
  Rochan H. Madhusudhana\\
  \affaddr{Georgia Institute of Technology}\\
  \email{rochan.hm@gatech.edu}
  \and
  Rahul K. Dass\\
  \affaddr{Georgia Institute of Technology}\\
  \email{rdass7@gatech.edu}
  \and
  Jeanette Luu\\
  \affaddr{Georgia Institute of Technology}\\
  \email{jeanette.luu@gatech.edu}
  \and
  Ashok K. Goel\\
  \affaddr{Georgia Institute of Technology}\\
  \email{ashok.goel@cc.gatech.edu}
}

\maketitle

\begin{sloppypar}
\begin{abstract}
In online learning, the ability to provide quick and accurate feedback to learners is crucial.
In skill-based learning, learners need to understand the underlying concepts and mechanisms of a skill to be able to apply it effectively.
While videos are a common tool in online learning, they cannot comprehend or assess the skills being taught.
Additionally, while Generative AI methods are effective in searching and retrieving answers from a text corpus, it remains unclear whether these methods exhibit any true understanding.
This limits their ability to provide explanations of skills or help with problem-solving.
This paper proposes a novel approach that merges Cognitive AI and Generative AI to address these challenges.
We employ a structured knowledge representation, the TMK (Task-Method-Knowledge) model, to encode skills taught in an online Knowledge-based AI course.
Leveraging techniques such as Large Language Models, Chain-of-Thought, and Iterative Refinement, we outline a framework for generating reasoned explanations in response to learners' questions about skills.
\end{abstract}


\section{Introduction}
Online education such as massive open online courses, online professional certifications and online academic degrees have empowered learners with greater accessibility to education across domains and disciplines.
With increased scalability and affordability, it is unsurprising that a significant portion of adult online learners are workers, most of whom require reskilling and upskilling \cite{WEF2019,goel2024ai}.
Videos in adult online education are a common source by which workers learn about new skills.
The term ``skill'' implies the procedures for accomplishing a task.
``Skill learning'', also known as procedural memory, is the capacity to acquire any ability through practice and repetition \cite{squire_skill_learning, jeremy_skill_learning}.
However, simply watching educational videos and answering quizzes fosters passive learning by learners. 
Based on literature from learning and education, active forms of learning often translates into a deeper understanding of the material taught and increased cognitive engagement by learners \cite{chi2018icap}.

Although there is a huge body of work that has investigated how students learn from a cognitive, education and learning sciences perspective \cite{chi2021}, there is scant literature that considers how the educational technology can not only capture a deep \textit{understanding} of the skills being taught but also have a capacity to address complex or novel learner inquiries.
By ``understanding'', in this context as stated in the learning and cognition literature, we mean an entity's (a human learner's or even an AI agent's) ability to draw on correct inferences and not draw on incorrect inferences about the thing that is understood \cite{schank2013scripts}.

Emergent advances in Generative AI such as large language models (LLMs) continue to demonstrate remarkable capabilities in various tasks, most notably question-answering based on large text corpora \cite{chatgpt_kbqa, rag_survey}.
However, studies have indicated that LLMs do not autonomously manifest any understanding of their assigned tasks \cite{valmeekam2023planning}.
Therefore, at a high-level, we ask \textit{``Do LLMs understand skills taught in educational settings?''}.

To address this gap, we propose a novel theory and application of learning that integrates Cognitive AI and Generative AI to enhance the representation and explanation of skills within online learning environments.
From a Cognitive AI perspective, we consider each skill taught in a course as an 'abstract device'.
As an abstract device, a skill has a design that includes specific functions and components, each with its own purpose.
The design also incorporates causal mechanisms that combine these component functions into a comprehensive understanding of the skill.
This approach emphasizes a hierarchical representation, where understanding a skill involves understanding its components, their interactions, and the teleological aspect—how these elements contribute to the overall purpose or goal of the skill.

Using a structured knowledge representation framework called TMK (Task-Method-Knowledge) \cite{murdock2008meta_redacted,rugaber2013gaia}, we conceptually capture problem-solving skills by decomposing it into three components: \textit{Task} pertains to the goal of the skill, \textit{Method} referring to the mechanism for accomplishing the goal, and \textit{Knowledge} specifying the objects and operations available in the environment. 
By leveraging Generative AI methods like GPT-based LLMs, Chain-of-Thought \cite{cot}, and Iterative Refinement \cite{selfrefine}, we develop an intelligent agent (IA) that generates contextually rich and reasoned explanations to learners' inquiries about skills.

In this paper, we postulate two research questions (RQs) and research hypotheses (RHs) to explore this convergence of cognitive AI and generative AI for our given context: 
\begin{description}
    \item[RQ1:] How may an IA explain how a skill works and explain the results of the skill in a specific instance?
    \begin{description}
        \item[RH1:]By inspecting on its design and simulating the execution of the skill and reflecting on the trace of the simulation on the instance.
    \end{description}
    \item[RQ2:]How may an IA inspect on the design of a skill and reflect on the trace of simulating the skill on an instance?
    \begin{description}
        \item[RH2:]By representing the skill's design as a TMK model and keeping a derivational trace of actual processing on the instance.
    \end{description}
\end{description}

\section{Related Work}
Based on the RQs and RHs from the previous section, we discuss related work revolving around two themes: (1) \textit{Representing skills in AI}, and (2) \textit{Generating answers or explanations using AI.}

\subsection{Representing Skills in AI}

Prior work instigating how to represent skills, particularly in domains such as AI in education and robotics have been made.
Well-established intelligent tutoring systems like Cognitive Tutors \cite{anderson1995cognitive,koedinger2006cognitive,rau2009intelligent} use rule-based formulations based on the ACT* (Architecture of Cognition) theory of learning and problem solving \cite{anderson1983architecture,anderson1993rules}.
In addition to representing cognitive skills (goal-oriented knowledge), these tutors have successfully enabled learners to acquire other skills in domains like programming skills in LISP \cite{anderson1984learning}, geometry \cite{anderson1981acquisition}, and fractions \cite{rau2009intelligent}.
Other approaches like an ontology-based framework for a structured representation of cognitive skills in K-12 settings have also been proposed \cite{cogskillnet}. 

But, how do humans represent knowledge?
More importantly, can we represent knowledge using a common framework in AI-based systems?
Chi et al. (1981) state that humans consider the knowledge in a given domain as \textit{world state representations} \cite{chi1981categorization}.
While domain experts view these representations to be comprised of ``deep functional features'', novices tend to only consider them as ``shallow perceptual features'' \cite{li2012efficient}.
Liu et al. (2012, 2015) \cite{li2012efficient,li2015integrating} integrated an expert representation (deep features) from the domain of algebra into SimStudent \cite{lee2009computational}, a machine learning agent to help student learn through examples and get feedback from a cognitive tutor.

In robotics, skill representation is a foundational problem for various applications like planning, task completion and human-robot interaction \cite{kroemer2021review,fitzgerald2021abstraction}.
Vassos and Christensen (2013) propose a framework combining planning with formal task modeling to automate task modeling and execution planning in manufacturing robotics \cite{6697194}.

Topp et al. (2018) \cite{8593566} state that representing skills using knowledge-based approaches such as ontologies is also helpful for end-user programming of synchronized motions between robot arms or between human-robot interaction.

\subsection{Generating Answers Using AI}
Several studies have investigated knowledge-based question answering (KBQA) using various AI techniques \cite{knowledge_repr_qa,moldovan2002lcc,chu2003question,tari2005using}. 
Braz et al. (2005) considers a hierarchical knowledge representation called EFDL (Extended Feature Description Logic) and uses Integer Linear Programming and phrase-level subsumption algorithms to generate answers to existing QA databases \cite{de2005knowledge}.
Balduccini et al. (2008) converts English text to logical representation and then uses automated logical theorem provers to extract facts and answer questions \cite{balduccini2008knowledge}.
Leveraging structured knowledge base like databases, Anette et al. (2007) proposed a hybrid NLP-based architecture consisting of question, analysis, search, extraction, and answer preparation stages in multiple lanaguages \cite{frank2007question}.

Given the recent capabilities of LLMs, they have also been used in KBQA studies.
Tan et al. (2023) leverages the few-shot learning capabilities of LLMs and proposes an in-context-learning framework to answer questions.
By using a rank-based KBQA method and representing the questions as multiple choice questions, they provide the LLM with in-prompt examples to improve the LLM's answers for complex questions.
Other rank-based approaches have also been used \cite{ye2021rng}.
ReTraCk \cite{chen2021retrack} is another LLM-based architecture for large-scale KBQA.
The authors propose to convert questions into a logical form and then uses SPARQL queries and a BERT NER model to retrieve answers.

Together, all these studies have shown that there is a potential of leveraging cognitive AI and Generative AI methods for various use cases.
However, very few studies consider modeling problem-solving skills as ``abstract devices'' that can be introspected for generating explanations to user questions within the context of online learning environments.

To address this gap in the literature, the next section describes a novel method that models a skill taught in an online course using a structured knowledge representation.
Leveraging generative AI methods, we illustrate how an Interactive Video system, or `Ivy', generates reasoned explanations in response to users’ questions about a skill.
While Ivy also enhances the interactivity of online educational videos, this paper focuses solely on the question-answering component of the framework.

\section{Methodology}

\begin{figure}[htbp]
  \centering
  \includegraphics[scale=0.75]{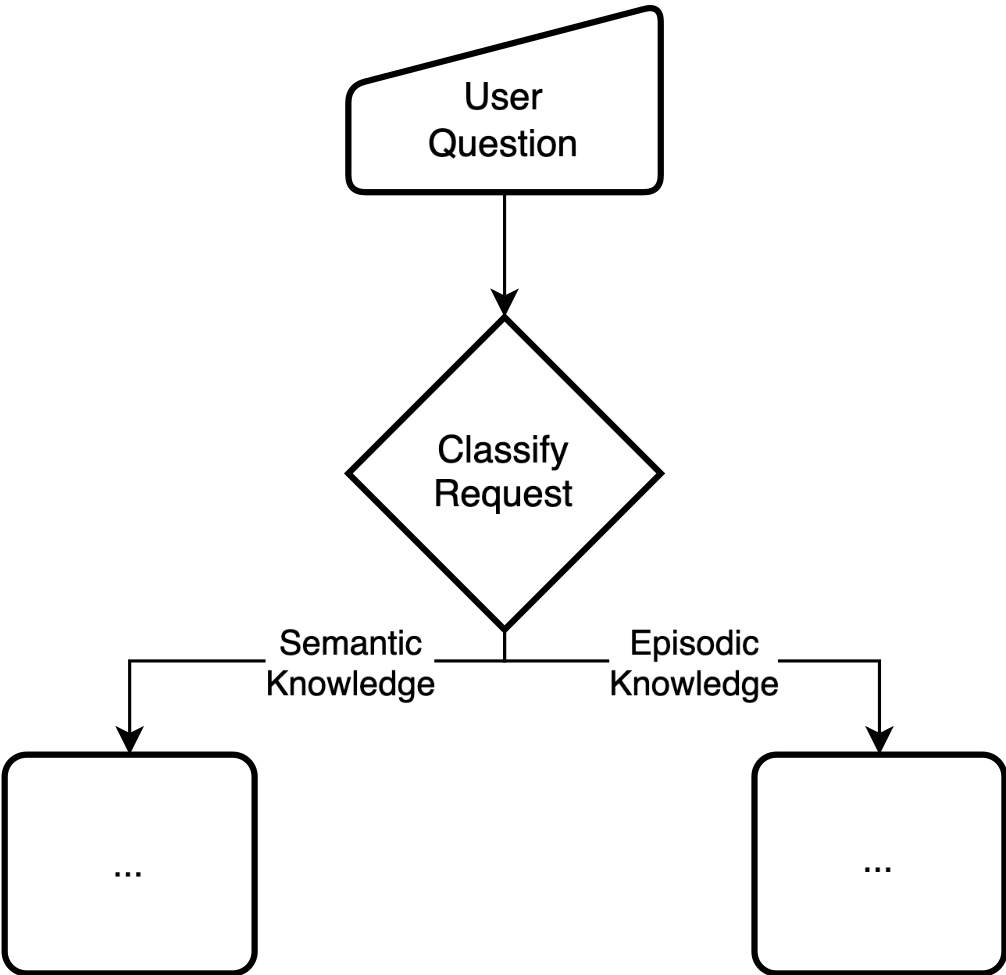}
  \caption{High-level flow of Ivy}
  \Description{High-level flow of Ivy}
  \label{fig:ivy-high-level-flow}
\end{figure}

In cognitive science, it is well understood that declarative memory consists of both semantic memory, which encompasses facts and concepts, and episodic memory, which includes events and experiences \cite{structural_episodic, declarative_memory}.
Our proposed ``Interactive Video'' system, or Ivy, draws inspiration from this cognitive science theory to answer questions.
When a user asks a question to the system, Ivy first classifies user questions as either related to semantic knowledge - pertaining to theories, frameworks, concepts, or definitions - or to episodic knowledge - which concerns events, experiences, or specific instances of a concept.

To classify questions, Ivy employs generative AI methods, specifically a Large Language Model (LLM) based on GPT-3.5 Turbo, capable of zero-shot classification.
This model uses in-context examples to determine whether a question relates to semantic or episodic knowledge \cite{fewshotllm}.
Typically, questions formatted as ``What is X?'' or ``Explain X'' are categorized under semantic knowledge, while questions like ``How do I do X?'' or ``What is the process to achieve X?'' fall under episodic knowledge.
The latter type of question is generally more challenging to answer as it often requires a simulation and an introspective analysis of the simulation process. This paper, however, focuses on questions related to semantic knowledge. 

\subsection{TMK Modeling}
\begin{figure}[htbp]
  \centering
  \includegraphics[scale=0.75]{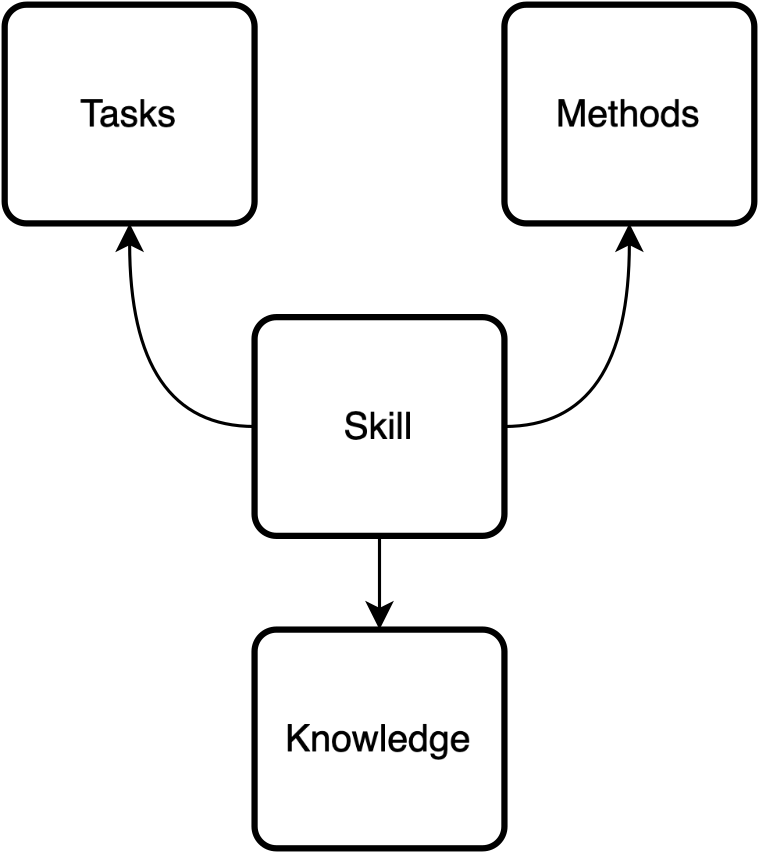}
  \caption{TMK Modeling in Ivy}
  \Description{TMK Modeling in Ivy}
  \label{fig:ivy-tmk}
\end{figure}

The TMK model is central to Ivy's functionality. TMK stands for ``Task-Method-Knowledge'', and is the primary knowledge representation model used by Ivy. The TMK model guides both the system's retrieval process and response generation. To integrate a new skill or concept into Ivy's capabilities, we first develop a TMK model of it, which involves several steps \cite{murdock2000semi_redacted}:

\begin{itemize}
  \item \textbf{Task Definition}: We identify the goal or purpose of a given skill. This involves specifying the allowable inputs (referred to as \texttt{givens}) and the expected outputs (\texttt{makes}). 
  For example, in a sorting algorithm, the goal could be to sort a list of numbers in ascending order.

  \item \textbf{Method Specification}: We describe the mechanisms for acomplishing the tasks. This is generally done using a deterministic finite state machine (FSM) called an \texttt{Organizer}. Each \texttt{Organizer} consists of a set of states, transitions, and actions that guide the system through the process of achieving the task. Each state may also have sub-goals within their own mechanisms, enabling hierarchical modeling. 
  For example, in a sorting algorithm, the methods could involve comparing pairs of numbers and swapping them if they are out of order.

  \item \textbf{Knowledge Representation}: We define the objects, concepts, and relationships within the environment. This includes properties of the objects and the logical expressions that connect user-supplied values. 
  Going back to the sorting algorithm example, the knowledge representation could include the concepts of numbers, lists, and the relationships between them.
\end{itemize}

\subsection{Question Classification}
\begin{figure}[htbp]
  \centering
  \includegraphics[scale=0.75]{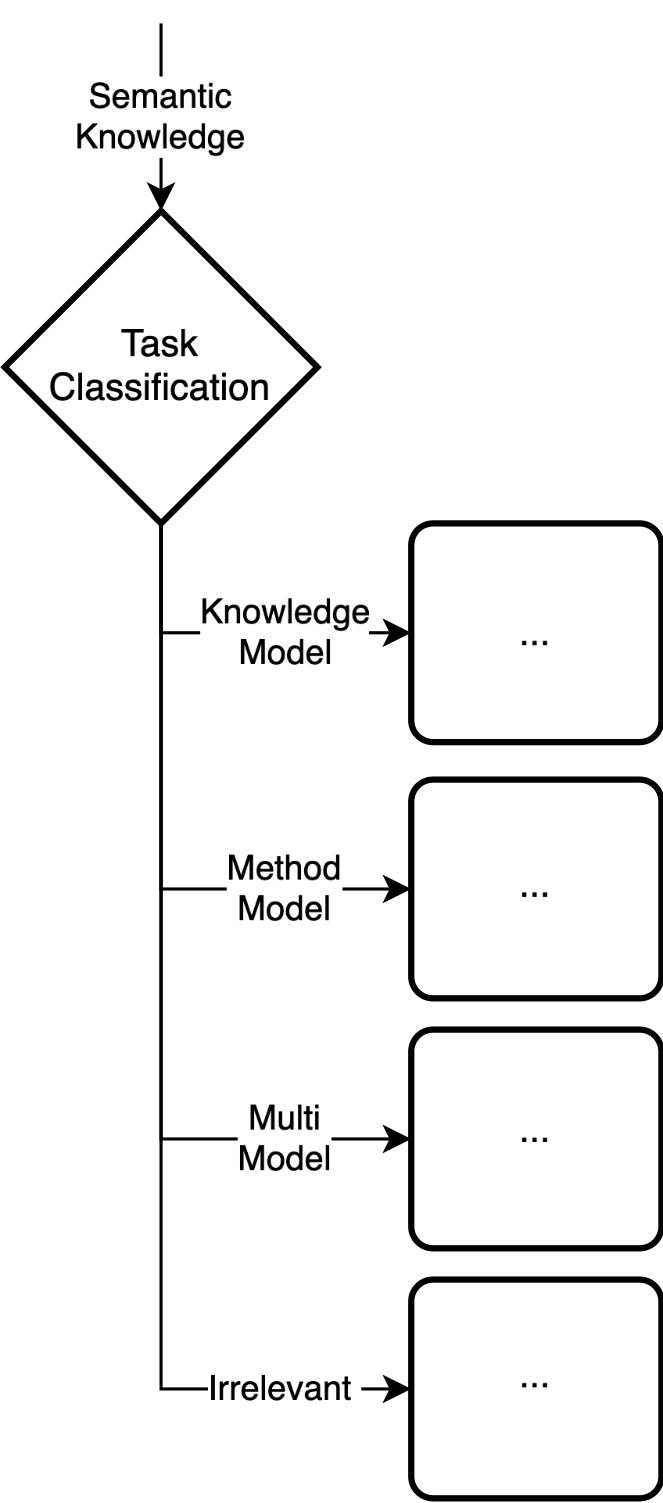}
  \caption{Semantic Knowledge Branch of Ivy}
  \Description{Semantic Knowledge Branch of Ivy}
  \label{fig:ivy-semantic-flow}
\end{figure}

Once a question has been categorized as pertaining to semantic knowledge, Ivy proceeds to classify it further based on the components of the TMK model. The system distinguishes whether the question relates to the Knowledge Model, the Method/Task Model, or the Multi Model. The Task and Method are interrelated within TMK, with the Task describing the goal of the skill and the Method outlining the mechanisms to achieve that goal; therefore, they are grouped together for classification. If a question does not pertain to the skill or concept being addressed, it is classified as Irrelevant. Ivy uses an LLM for zero-shot classification, providing it with descriptions and examples of each category. The model then classifies the question into one of these categories or as Irrelevant if it does not relate to the taught skill or concept.

\begin{itemize}
\item \textbf{Knowledge Model}: Questions related to the objects, concepts, and relationships within the environment. These questions focus on the properties of objects and the logical connections between user-supplied values. For example, in a sorting algorithm, a Knowledge Model question might ask about the properties of the numbers being sorted.

\item \textbf{Method/Task Model}: Questions concerning the mechanisms for accomplishing tasks. These questions address the states, transitions, and actions that guide the system conceptually through the process of achieving the task, without executing it. For example, in a sorting algorithm, a Method Model question might ask about the process of comparing pairs of numbers and swapping them if they are out of order, without actually running the sorting algorithm.

\item \textbf{Multi Model}: Questions that involve both the Knowledge Model and Method/Task Model. These questions explore the interplay between the objects, concepts, and relationships within the environment and the mechanisms for completing tasks. For example, in a sorting algorithm, a Multi Model question might ask about how the properties of numbers influence the process of comparing and sorting them.
\end{itemize}

\subsection{Relevant Knowledge Retrieval}
\begin{figure}[htbp]
  \centering
  \includegraphics[scale=0.75]{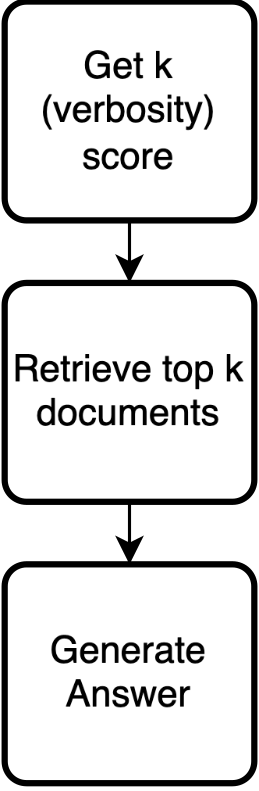}
  \caption{Knowledge Retrieval Flow in Ivy}
  \Description{Knowledge Retrieval Flow in Ivy}
  \label{fig:ivy-knowledge-retrieval-flow}
\end{figure}

Once the question has been classified as a Knowledge Model, Method/Task Model, or Multi Model question, Ivy retrieves the most relevant documents from the TMK knowledge base. The process of retrieving relevant documents involves two key steps:

\subsubsection{K-Score Calculation}
Ivy calculates a \texttt{k-score} ranging from 1 to 4 based on the verbosity of the expected response. The k-score helps determine the level of detail and complexity in the response. For instance, a k-score of 1 suggests a very brief response (3-5 words), while a score of 2 indicates a short response of a few sentences. A k-score of 3 corresponds to a more detailed paragraph-length response, and a k-score of 4 leads to a comprehensive response spanning multiple paragraphs.

\subsubsection{Document Retrieval}
Depending on the \texttt{k-score}, Ivy retrieves the top $k$ relevant documents from the TMK knowledge base. This retrieval is facilitated by the FAISS library \cite{faiss}, which is designed for efficient similarity search and clustering of dense vectors, making it ideal for finding relevant documents based on the input question. The documents are ranked according to their relevance to the question, allowing Ivy to focus on the most pertinent information for response generation.

For questions classified as Knowledge Model, only the top $k$ Knowledge documents are retrieved. Similarly, for questions under the Multi Model category, the top $k$ documents from a combined set of Knowledge, Task, and Method documents are retrieved.
In contrast, for questions classified as Method/Task Model, the retrieval focuses on the most relevant Task Document and its associated Method Document.

\subsection{Response Generation}

\begin{figure}[htbp]
  \centering
  \includegraphics[scale=0.75]{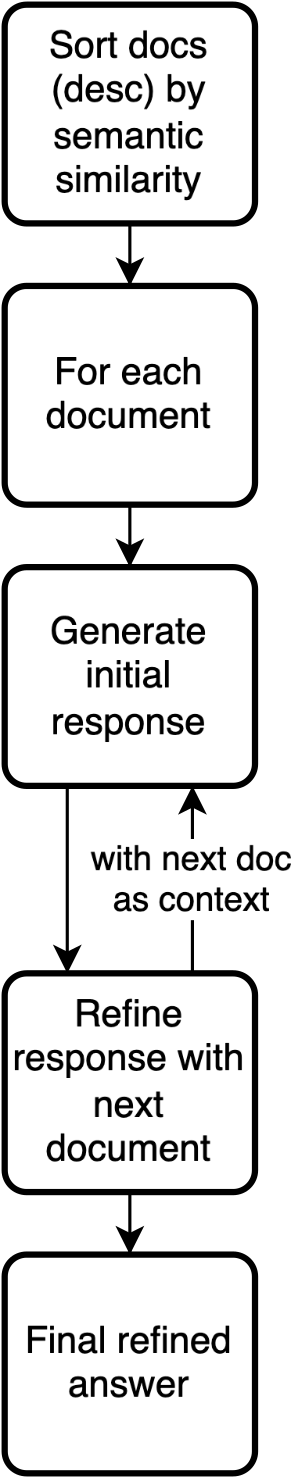}
  \caption{Iterative Refinement Flow in Ivy}
  \Description{Iterative Refinement Flow in Ivy}
  \label{fig:ivy-refinement-flow}
\end{figure}

Once the relevant documents have been retrieved, Ivy generates a response based on the classification of the question and the retrieved documents. The response generation process involves one of the following steps:

\subsubsection{Iterative Refinement}
For questions classified as Knowledge Model or Multi Model, Ivy employs an iterative refinement process to generate a response. The iterative refinement process involves the following steps:

\begin{itemize}
\item \textbf{Initial Response Generation}: Ivy generates an initial response based on the top most relevant document retrieved from the TMK knowledge base. The initial response is generated solely using the information contained in the document.
\item \textbf{Refinement}: Using the remaining $k-1$ relevant documents, Ivy refines the initial response by incorporating additional information and context. The refinement process aims to enhance the response's accuracy, completeness, and relevance by iteratively incorporating relevant details from the retrieved documents.
\end{itemize}

Each step in the iterative refinement process involves a call to the LLM to generate natural language responses. For the initial response generation step, a call is made to the LLM with the user query, the expected verbosity level (determined by the \texttt{k-score}), and the top most relevant document, provided as context. For the refinement step, the LLM is called with the user query, the initial response generated in the previous step, and the remaining relevant documents. The LLM synthesizes the information from the documents to refine the initial response and generate a more comprehensive answer. 

\subsubsection{Chain-of-Thought Generation}
For questions classified as Method/Task Model, Ivy employs a Chain-of-Thought generation process to generate a response. The process involves two key steps:

\begin{itemize}
\item \textbf{Transition Extraction}: Ivy extracts all the relevant transitions from the topmost Method document. These transitions are part of the finite state machine (FSM) that guides the system through the process of achieving the task. Each transition represents a step or action in the task completion process, and includes the associated states, actions, and conditions.
\item \textbf{Transition Explanation Generation}: Using the extracted transitions and the metadata, Ivy generates a natural language response that explains the task completion process through the transitions, ensuring the conditions, actions, and states are clearly articulated. The response provides a detailed explanation of the method and the steps involved in achieving the task.
\end{itemize}

The Transition Explanation Generation step leverages the information contained in the transitions to generate a coherent and detailed response that elucidates the method and task completion process. Chain-of-Thought prompting \cite{cot} is used to guide the generation process, with multiple in-prompt examples provided to the LLM to demonstrate the reasoning and context required for stitching together the transitions (which might be in a non-linear order). Further, the expected verbosity level (determined by the \texttt{k-score}) is provided to guide the response generation length.

\section{A Worked Example}

To better illustrate the application of our approach, we use the classic river crossing problem from an AI course as an example. This problem involves three missionaries and three cannibals, or alternatively, three guards and three prisoners, who must cross a river using a boat that can carry at most two people. If the number of prisoners on either side of the river exceeds the number of guards, the prisoners will overpower the guards and escape. The goal is to transport all individuals to the other side of the river without violating this rule.

When a user poses the question, ``Who is a guard?'', the following steps are taken to generate a response:

\subsection{Step 1: TMK Modeling}

We begin by creating a highly simplified TMK model for the river crossing problem.

\begin{figure}[htbp]
  \centering
  \includegraphics[scale=0.75, width=\columnwidth]{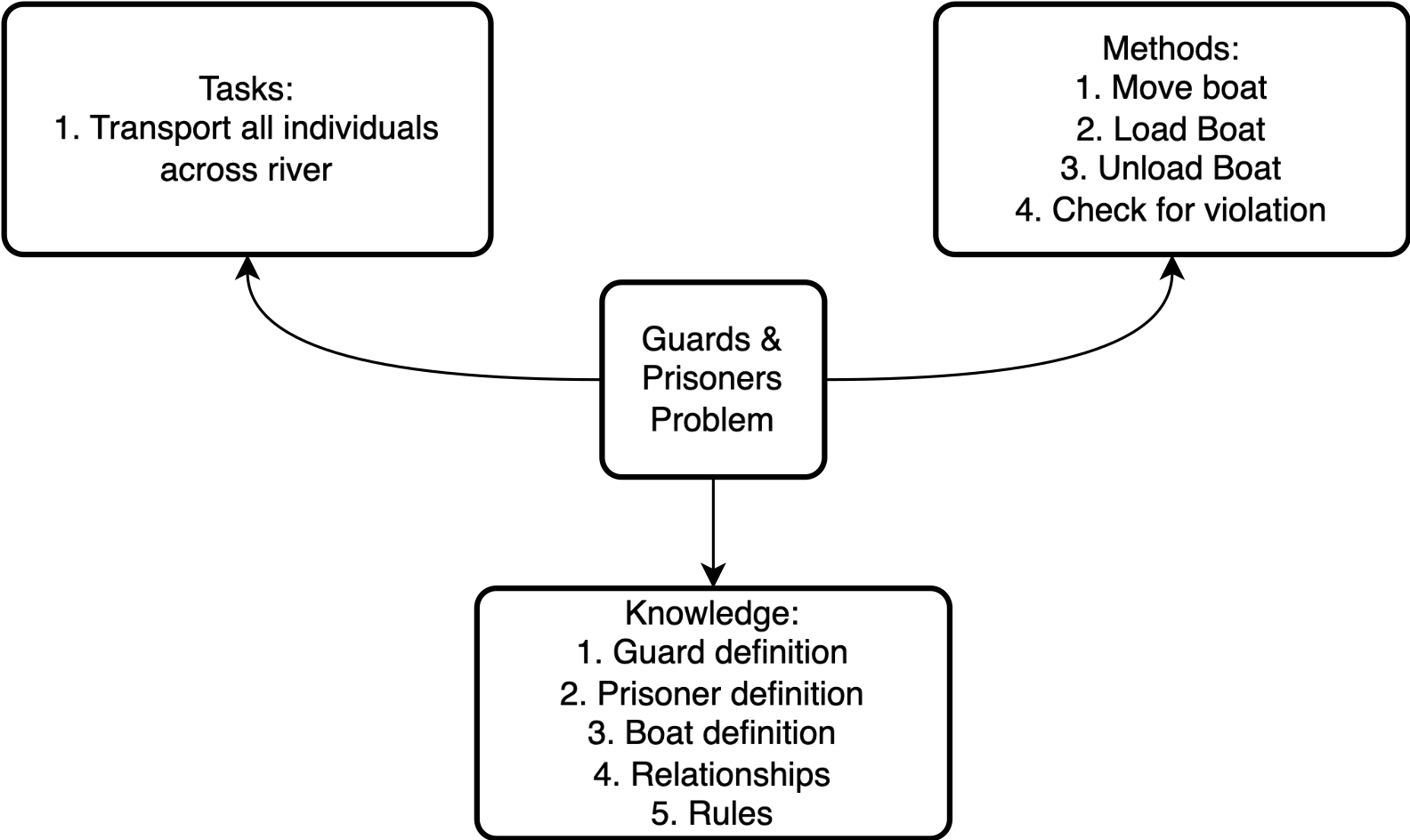}
  \caption{TMK Model for River Crossing Problem}
  \Description{TMK Model for River Crossing Problem}
  \label{fig:ivy-gpp-tmk}
\end{figure}

Here, our primary task might be defined as \texttt{Transport All Individuals Across the River}. The method could involve steps such as: \texttt{Move Boat to Opposite Bank}, \texttt{Load Boat with Selected Individuals}, \texttt{Unload Boat}, and \texttt{Check for Constraint Violation}. The knowledge representation might include the concepts of \texttt{Guards}, \texttt{Prisoners}, \texttt{Boat}, the relationships between them, and the rules governing the movement of individuals.

\subsection{Step 2: Question Classification}

To classify the question "Who is a guard?", we submit a query to the LLM that includes a description of the TMK model components and a prompt for classification:

\vspace{-1em}
\begin{verbatim}
Classify the following question as either a 
Knowledge Model, Method/Task Model, Multi Model, 
or Irrelevant question:

Question: Who is a guard?
Context:
    Knowledge Model: 
    <Description of Knowledge Model>
    In the context of this problem,  we have the 
    following knowledge entities:
        Guards: <Description of Guards>
        Prisoners: <Description of Prisoners>
        ...
    Method/Task Model:
    <Description of Method/Task Model>
    In the context of this problem, we have the 
    following tasks:
        Transport All Individuals Across the 
        River: <Description of Task and its 
                methods>
        ...
    Multi Model:
    <Description of Multi Model>
    In the context of this problem, we have the 
    following Task, Method, and Knowledge entities:
    ...
    Irrelevant:
    <Description of Irrelevant>
    Here are some example questions, TMK models,
    and classifications:
    ...
\end{verbatim}
\vspace{-1em}

The LLM would likely classify the question as a Knowledge Model question because it pertains to understanding the entities within the environment.

\subsection{Step 3: Relevant Knowledge Retrieval}

After classifying the question, we proceed to retrieve the relevant Knowledge documents from the TMK knowledge base. The first step is calculating the \texttt{k-score} based on the verbosity expected in the response. For a question like ``Who is a guard?'', a k-score of 2 might be appropriate, indicating a short answer. This helps in determining the number of relevant documents to retrieve.

We then use the FAISS library to perform a similarity search between the user query and all Knowledge documents in the TMK knowledge base. The top $k = 2$ documents with the highest similarity scores are retrieved for response generation. These documents might be:

\begin{itemize}
  \item Document 1 - \texttt{Guard Definition}: Contains the definition and description of the guards in the river crossing problem. Similarity score: 0.60
  \item Document 2 - \texttt{Relationships}: Contains information about the relationships between guards and prisoners in the river crossing problem. Similarity score: 0.45
\end{itemize}

\subsection{Step 4: Response Generation}

As the question ``Who is a guard?'' is classified as a Knowledge Model question, we employ the iterative refinement process to generate a response. An initial request of the following form is made to the LLM with the user query and the top most relevant document (Document 1) to generate an initial response:

\vspace{-1em}
\begin{verbatim}
Generate a response to the following question:

Question: Who is a guard?
Context: <Content of Document 1>
\end{verbatim}
\vspace{-1em}

The LLM generates an initial response based on the user query and the content of Document 1. The initial response might be: ``In the river crossing problem, the guards are one of the individuals who need to be transported across the river.'' This response is then further refined by incorporating additional information from Document 2. A request of the following form is made to the LLM for refinement:

\vspace{-1em}
\begin{verbatim}
Refine the response to the following question 
if necessary:

Question: Who is a guard?
Context: 
    Initial Response: In the river crossing 
    problem, the guards are individuals 
    who need to be transported across 
    the river.
Additional Context: <Content of Document 2>
\end{verbatim}
\vspace{-1em}

The LLM synthesizes the information from Document 2 to refine the initial response and generate a more comprehensive answer. The refined response might be: ``In the river crossing problem, the guards are individuals who need to be transported across the river. They play a crucial role in ensuring that the prisoners do not escape during the crossing.''

This worked example demonstrates how our system uses the TMK model to structure its knowledge base and respond effectively to user queries.


\begin{table*}[ht!]
    \centering
    \resizebox{\textwidth}{!}{%
    \begin{tabular}{|p{0.2\linewidth}|p{0.4\linewidth}|p{0.4\linewidth}|}
        \hline
        \textbf{Metric} & \textbf{Definition} & \textbf{Scale Description} \\
        \hline
        Completeness & Measures how much of the relevant information the response covers compared to a reference answer (TMK Document). & 
        \parbox[t]{\linewidth}{\raggedright 1: Very Incomplete (less than 20\% of relevant information) \\ 2: Incomplete (20-40\%) \\ 3: Moderately Complete (40-60\%) \\ 4: Complete (60-80\%) \\ 5: Very Complete (80-100\%)} \\
        \hline
        Precision & Assesses the accuracy and relevance of the information provided in the response. & 
        \parbox[t]{\linewidth}{\raggedright 1: Very Low Precision (many inaccuracies or irrelevant details) \\ 2: Low Precision \\ 3: Moderate Precision \\ 4: High Precision \\ 5: Very High Precision (almost all information is accurate and relevant)} \\
        \hline
        Consistency & Evaluates the uniformity of the responses for the same query over multiple runs. & 
        \parbox[t]{\linewidth}{\raggedright 1: Very Inconsistent \\ 2: Inconsistent \\ 3: Moderately Consistent \\ 4: Consistent \\ 5: Very Consistent (responses are the same across all runs)} \\
        \hline
    \end{tabular}%
    }
    \caption{Proposed Evaluation Metrics}
    \label{tab:eval_metrics}
\end{table*}

\section{Conclusion}

In this paper, we have presented an innovative integration of Cognitive AI with Generative AI to enhance question answering in skill-based learning environments. Our approach utilizes the Task-Method-Knowledge (TMK) model to systematically encode the skills being taught, providing a structured and detailed representation that supports accurate and contextually rich explanations. 

Combining the TMK model with Generative AI techniques, including Large Language Models (LLMs), Chain-of-Thought, and Iterative Refinement, we have developed a system, Ivy, that can produce answers that are not only contextually relevant but also pedagogically relevant. The TMK model captures the hierarchical structure of skills, enabling Ivy to generate reasoned explanations. We envision that this approach might be particularly beneficial in online education, where dynamic and interactive content is essential for effective learning. The goal of such systems would be to bridge the gap between passive content consumption and active learning engagement by simulating a deeper understanding and providing interactive feedback to learners.

The focus of the system described above is on semantic knowledge question answering. However, the TMK framework also holds potential for expanding into episodic knowledge question answering. By using the \texttt{Organizer} components in the Method specification of a TMK model, Ivy can simulate problem-solving processes and generate logs of the states visited and transitions taken during the problem-solving process. By inspecting these logs, Ivy can generate detailed explanations of the problem-solving process, allowing it to answer ``What if?'' and ``How?'' questions that pertain to specific instances or events.

\subsection{Future Work}
Looking ahead, several avenues for future work and improvement exist. 

\begin{enumerate}
  \item \textbf{Evaluation Mechanism}: We aim to refine and standardize the evaluation mechanism for assessing performance of our TMK-based system. The evaluation metrics outlined in Table \ref{tab:eval_metrics} - Completeness, Precision and Consistency - provide an overview of the system's performance, but further refinement and validation might be needed to ensure robust evaluation. The current metrics are based on an AI engineer's subjective evaluation (using a 5-point Likert scale \cite{likert}). However, we plan to implement a mixed-methods evaluation approach that combines automated metrics (e.g., BLEU, ROUGE) with human evaluations to measure the output quality more effectively.
  \item \textbf{Automation of TMK Model Creation}: Currently, the creation of the TMK model is a manual process that requires domain expertise. Automating the development of TMK models can streamline the setup process for new educational content, making it more scalable and adaptable to diverse learning domains.
  \item \textbf{Enhanced Generative AI Techniques}: Further refining the integration of Chain-of-Thought and Iterative Refinement techniques may improve the system's ability to handle more nuanced and complex questions, enhancing the quality and depth of responses generated.
\end{enumerate}

\section{Acknowledgments}
We are grateful to Spencer Rugaber at Georgia Tech's Design Intelligence Laboratory for his invaluable insights into TMK models and modeling.
We thank Cherie Lum, Erin Deye, and Sashank Verma for many discussions about this work
and critiques of earlier drafts of this paper.
This research has been supported by NSF Grants \#2112532 and \#2247790 awarded to the National AI Institute for Adult Learning and Online Education.

%

\bibliography{main}  

\begin{thebibliography}{10}

\bibitem{WEF2019}
H.~Leurent, F.~Betti, E.~Shook, R.~Fuchs, and F.~Damrath, ``Leading through the fourth industrial revolution: putting people at the centre,'' in {\em World Economic Forum}, pp.~1--25, 2019.

\bibitem{goel2024ai}
A.~Goel, C.~Dede, M.~Garn, and C.~Ou, ``Ai-aloe: Ai for reskilling, upskilling, and workforce development,'' {\em AI Magazine}, 2024.

\bibitem{squire_skill_learning}
L.~R. Squire, ``{Declarative and Nondeclarative Memory: Multiple Brain Systems Supporting Learning and Memory},'' {\em Journal of Cognitive Neuroscience}, vol.~4, pp.~232--243, 07 1992.

\bibitem{jeremy_skill_learning}
J.~Doyon, ``Skill learning,'' in {\em International Review of Neurobiology} (J.~D. Schmahmann, ed.), vol.~41 of {\em International Review of Neurobiology}, pp.~273--294, Academic Press, 1997.

\bibitem{chi2018icap}
M.~T. Chi, J.~Adams, E.~B. Bogusch, C.~Bruchok, S.~Kang, M.~Lancaster, R.~Levy, N.~Li, K.~L. McEldoon, G.~S. Stump, {\em et~al.}, ``Translating the icap theory of cognitive engagement into practice,'' {\em Cognitive science}, vol.~42, no.~6, pp.~1777--1832, 2018.

\bibitem{chi2021}
M.~T.~H. Chi, ``Translating a theory of active learning: An attempt to close the research-practice gap in education,'' {\em Topics in Cognitive Science}, vol.~13, no.~3, pp.~441--463, 2021.

\bibitem{schank2013scripts}
R.~C. Schank and R.~P. Abelson, {\em Scripts, plans, goals, and understanding: An inquiry into human knowledge structures}.
\newblock Psychology press, 2013.

\bibitem{chatgpt_kbqa}
Y.~Tan, D.~Min, Y.~Li, W.~Li, N.~Hu, Y.~Chen, and G.~Qi, ``Can chatgpt replace traditional kbqa models? an in-depth analysis of the question answering performance of the gpt llm family,'' in {\em The Semantic Web -- ISWC 2023} (T.~R. Payne, V.~Presutti, G.~Qi, M.~Poveda-Villal{\'o}n, G.~Stoilos, L.~Hollink, Z.~Kaoudi, G.~Cheng, and J.~Li, eds.), (Cham), pp.~348--367, Springer Nature Switzerland, 2023.

\bibitem{rag_survey}
Y.~Gao, Y.~Xiong, X.~Gao, K.~Jia, J.~Pan, Y.~Bi, Y.~Dai, J.~Sun, M.~Wang, and H.~Wang, ``Retrieval-augmented generation for large language models: A survey,'' 2024.

\bibitem{valmeekam2023planning}
K.~Valmeekam, M.~Marquez, S.~Sreedharan, and S.~Kambhampati, ``On the planning abilities of large language models-a critical investigation,'' {\em Advances in Neural Information Processing Systems}, vol.~36, pp.~75993--76005, 2023.

\bibitem{murdock2008meta_redacted}
Anonymous, ``Redacted for blind review,'' {\em Redacted}, vol.~20, no.~1, pp.~1--36, 2008.

\bibitem{rugaber2013gaia}
S.~Rugaber, A.~K. Goel, and L.~Martie, ``Gaia: A cad environment for model-based adaptation of game-playing software agents,'' {\em Procedia Computer Science}, vol.~16, pp.~29--38, 2013.

\bibitem{cot}
J.~Wei, X.~Wang, D.~Schuurmans, M.~Bosma, B.~Ichter, F.~Xia, E.~Chi, Q.~Le, and D.~Zhou, ``Chain-of-thought prompting elicits reasoning in large language models,'' 2023.

\bibitem{selfrefine}
A.~Madaan, N.~Tandon, P.~Gupta, S.~Hallinan, L.~Gao, S.~Wiegreffe, U.~Alon, N.~Dziri, S.~Prabhumoye, Y.~Yang, S.~Gupta, B.~P. Majumder, K.~Hermann, S.~Welleck, A.~Yazdanbakhsh, and P.~Clark, ``Self-refine: Iterative refinement with self-feedback,'' 2023.

\bibitem{anderson1995cognitive}
J.~R. Anderson, A.~T. Corbett, K.~R. Koedinger, and R.~Pelletier, ``Cognitive tutors: Lessons learned,'' {\em The journal of the learning sciences}, vol.~4, no.~2, pp.~167--207, 1995.

\bibitem{koedinger2006cognitive}
K.~R. Koedinger, A.~Corbett, {\em et~al.}, {\em Cognitive tutors: Technology bringing learning sciences to the classroom}.
\newblock na, 2006.

\bibitem{rau2009intelligent}
M.~A. Rau, V.~Aleven, and N.~Rummel, ``Intelligent tutoring systems with multiple representations and self-explanation prompts support learning of fractions.,'' in {\em AIED}, pp.~441--448, 2009.

\bibitem{anderson1983architecture}
J.~R. Anderson, {\em The architecture of cognition}.
\newblock Harvard University Press, 1983.

\bibitem{anderson1993rules}
J.~R. Anderson, {\em Rules of the Mind}.
\newblock Lawrence Erlbaum Associates, Inc, 1993.

\bibitem{anderson1984learning}
J.~R. Anderson, R.~Farrell, and R.~Sauers, ``Learning to program in lisp,'' {\em Cognitive Science}, vol.~8, no.~2, pp.~87--129, 1984.

\bibitem{anderson1981acquisition}
J.~R. Anderson, J.~G. Greeno, P.~J. Kline, and D.~M. Neves, ``Acquisition of problem-solving,'' {\em Cognitive Skills and Their Acquisition}, vol.~16, p.~191, 1981.

\bibitem{cogskillnet}
P.~Askar and A.~Altun, ``Cogskillnet: An ontology-based representation of cognitive skills,'' {\em Journal of Educational Technology \& Society}, vol.~12, no.~2, pp.~240--253, 2009.

\bibitem{chi1981categorization}
M.~T. Chi, P.~J. Feltovich, and R.~Glaser, ``Categorization and representation of physics problems by experts and novices,'' {\em Cognitive science}, vol.~5, no.~2, pp.~121--152, 1981.

\bibitem{li2012efficient}
N.~Li, W.~W. Cohen, and K.~R. Koedinger, ``Efficient cross-domain learning of complex skills,'' in {\em Intelligent Tutoring Systems: 11th International Conference, ITS 2012, Chania, Crete, Greece, June 14-18, 2012. Proceedings 11}, pp.~493--498, Springer, 2012.

\bibitem{li2015integrating}
N.~Li, N.~Matsuda, W.~W. Cohen, and K.~R. Koedinger, ``Integrating representation learning and skill learning in a human-like intelligent agent,'' {\em Artificial Intelligence}, vol.~219, pp.~67--91, 2015.

\bibitem{lee2009computational}
A.~Lee, W.~W. Cohen, and K.~R. Koedinger, ``A computational model of how learner errors arise from weak prior knowledge,'' in {\em Proceedings of the Annual Conference of the Cognitive Science Society, Austin, TX}, pp.~1288--1293, 2009.

\bibitem{kroemer2021review}
O.~Kroemer, S.~Niekum, and G.~Konidaris, ``A review of robot learning for manipulation: Challenges, representations, and algorithms,'' {\em Journal of machine learning research}, vol.~22, no.~30, pp.~1--82, 2021.

\bibitem{fitzgerald2021abstraction}
T.~Fitzgerald, A.~Goel, and A.~Thomaz, ``Abstraction in data-sparse task transfer,'' {\em Artificial Intelligence}, vol.~300, p.~103551, 2021.

\bibitem{6697194}
J.~Huckaby, S.~Vassos, and H.~I. Christensen, ``Planning with a task modeling framework in manufacturing robotics,'' in {\em 2013 IEEE/RSJ International Conference on Intelligent Robots and Systems}, 2013.

\bibitem{8593566}
E.~A. Topp, M.~Stenmark, A.~Ganslandt, A.~Svensson, M.~Haage, and J.~Malec, ``Ontology-based knowledge representation for increased skill reusability in industrial robots,'' in {\em 2018 IEEE/RSJ International Conference on Intelligent Robots and Systems (IROS)}, 2018.

\bibitem{knowledge_repr_qa}
M.~Balduccini, C.~Baral, and Y.~Lierler, ``Chapter 20 knowledge representation and question answering,'' in {\em Handbook of Knowledge Representation} (F.~{van Harmelen}, V.~Lifschitz, and B.~Porter, eds.), vol.~3 of {\em Foundations of Artificial Intelligence}, pp.~779--819, Elsevier, 2008.

\bibitem{moldovan2002lcc}
D.~I. Moldovan, S.~M. Harabagiu, R.~Girju, P.~Morarescu, V.~F. Lacatusu, A.~Novischi, A.~Badulescu, and O.~Bolohan, ``Lcc tools for question answering.,'' in {\em TREC}, 2002.

\bibitem{chu2003question}
J.~Chu-Carroll, K.~Czuba, J.~Prager, and A.~Ittycheriah, ``In question answering, two heads are better than one,'' in {\em Proceedings of the 2003 Human Language Technology Conference of the North American Chapter of the Association for Computational Linguistics}, pp.~24--31, 2003.

\bibitem{tari2005using}
L.~Tari and C.~Baral, ``Using ansprolog with link grammar and wordnet for qa with deep reasoning,'' in {\em AAAI Spring Symposium Workshop on Inference for Textual Question Answering}, 2005.

\bibitem{de2005knowledge}
R.~de~Salvo~Braz, R.~Girju, V.~Punyakanok, D.~Roth, and M.~Sammons, ``Knowledge representation for semantic entailment and question-answering,'' in {\em IJCAI-05 workshop on knowledge and reasoning for answering questions}, pp.~71--80, 2005.

\bibitem{balduccini2008knowledge}
M.~Balduccini, C.~Baral, and Y.~Lierler, ``Knowledge representation and question answering,'' {\em Foundations of Artificial Intelligence}, vol.~3, pp.~779--819, 2008.

\bibitem{frank2007question}
A.~Frank, H.-U. Krieger, F.~Xu, H.~Uszkoreit, B.~Crysmann, B.~J{\"o}rg, and U.~Sch{\"a}fer, ``Question answering from structured knowledge sources,'' {\em Journal of Applied Logic}, vol.~5, no.~1, pp.~20--48, 2007.

\bibitem{ye2021rng}
X.~Ye, S.~Yavuz, K.~Hashimoto, Y.~Zhou, and C.~Xiong, ``Rng-kbqa: Generation augmented iterative ranking for knowledge base question answering,'' {\em arXiv preprint arXiv:2109.08678}, 2021.

\bibitem{chen2021retrack}
S.~Chen, Q.~Liu, Z.~Yu, C.-Y. Lin, J.-G. Lou, and F.~Jiang, ``Retrack: A flexible and efficient framework for knowledge base question answering,'' in {\em Proceedings of the 59th annual meeting of the association for computational linguistics and the 11th international joint conference on natural language processing: system demonstrations}, pp.~325--336, 2021.

\bibitem{structural_episodic}
D.~L. GREENBERG and M.~VERFAELLIE, ``Interdependence of episodic and semantic memory: Evidence from neuropsychology,'' {\em Journal of the International Neuropsychological Society}, vol.~16, p.~748–753, Jun 2010.

\bibitem{declarative_memory}
W.~J. Riedel and A.~Blokland, ``Declarative memory,'' {\em Cognitive Enhancement}, p.~215–236, 2015.

\bibitem{fewshotllm}
T.~B. Brown, B.~Mann, N.~Ryder, M.~Subbiah, J.~Kaplan, P.~Dhariwal, A.~Neelakantan, P.~Shyam, G.~Sastry, A.~Askell, S.~Agarwal, A.~Herbert-Voss, G.~Krueger, T.~Henighan, R.~Child, A.~Ramesh, D.~M. Ziegler, J.~Wu, C.~Winter, C.~Hesse, M.~Chen, E.~Sigler, M.~Litwin, S.~Gray, B.~Chess, J.~Clark, C.~Berner, S.~McCandlish, A.~Radford, I.~Sutskever, and D.~Amodei, ``Language models are few-shot learners,'' 2020.

\bibitem{murdock2000semi_redacted}
Anonymous, ``Redacted for blind review,'' {\em Redacted}, 2000.

\bibitem{faiss}
M.~Douze, A.~Guzhva, C.~Deng, J.~Johnson, G.~Szilvasy, P.-E. Mazaré, M.~Lomeli, L.~Hosseini, and H.~Jégou, ``The faiss library,'' 2024.

\bibitem{likert}
R.~Likert, ``A technique for the measurement of attitudes,'' {\em Archives of Psychology}, vol.~22, no.~140, p.~55, 1932.

\end{thebibliography}
%

\end{sloppypar}
\end{document}